\documentclass{article}
\usepackage{ijcai24}
\usepackage{graphbox}
\usepackage{times}
\usepackage{soul}
\usepackage{url}
\usepackage[utf8]{inputenc}
\usepackage[small]{caption}
\usepackage{graphicx}
\usepackage{amsmath}
\usepackage{amsthm}
\usepackage{booktabs}
\usepackage{algorithm}
\usepackage{amssymb}
\usepackage{algorithmic}
\usepackage{subfigure}
\usepackage{color}
\usepackage{graphicx}
\usepackage{multirow}
\usepackage[utf8]{inputenc}
\usepackage[colorlinks=true]{hyperref}
\usepackage{multirow}
\usepackage[dvipsnames, svgnames, x11names]{xcolor} 

\usepackage[switch]{lineno}

\hypersetup{
    linkcolor=red,
    citecolor=NavyBlue, % 设置引用文献的颜色
    urlcolor=black,
}

\title{Learning A Spiking Neural Network for Efficient Image Deraining}

\author{
Tianyu Song$^1$
\and
Guiyue Jin$^1$\and
Pengpeng Li$^2$\and
Kui Jiang$^3$\and
Xiang Chen$^2$\And
Jiyu Jin$^{1,*}$\\
\affiliations
$^1$Dalian Polytechnic University\\
$^2$Nanjing University of Science and Technology\\
$^3$Harbin Institute of Technology\\
\emails
songtienyu@163.com,
guiyue.jin@dlpu.edu.cn,
pengpengli@njust.edu.cn,
kuijiang\_1994@163.com,
chenxiang@njust.edu.cn,
jiyu.jin@dlpu.edu.cn
}
\begin{document}
\maketitle
\footnotetext{*Corresponding author.}
\begin{abstract}
Recently, spiking neural networks (SNNs) have demonstrated substantial potential in computer vision tasks.
In this paper, we present an \textbf{E}fficient \textbf{S}piking \textbf{D}eraining \textbf{Net}work, called ESDNet. 
Our work is motivated by the observation that rain pixel values will lead to a more pronounced intensity of spike signals in SNNs.
However, directly applying deep SNNs to image deraining task still remains a significant challenge.
This is attributed to the information loss and training difficulties that arise from discrete binary activation and complex spatio-temporal dynamics.
To this end, we develop a spiking residual block to convert the input into spike signals, then adaptively optimize the membrane potential by introducing attention weights to adjust spike responses in a data-driven manner, alleviating information loss caused by discrete binary activation.
By this way, our ESDNet can effectively detect and analyze the characteristics of rain streaks by learning their fluctuations. This also enables better guidance for the deraining process and facilitates high-quality image reconstruction.
Instead of relying on the ANN-SNN conversion strategy, we introduce a gradient proxy strategy to directly train the model for overcoming the challenge of training. 
Experimental results show that our approach gains comparable performance against ANN-based methods while reducing energy consumption by 54\%. 
The code source is available at~\url{https://github.com/MingTian99/ESDNet}.
\end{abstract}

%\vspace{-5mm}
\section{Introduction}
Single image deraining aims to reconstruct high-quality rain-free images from rain-affected ones.
This challenging problem has witnessed significant advances due to the development of various effective image priors~\cite{luo2015DSCg,li2016GMM} and deep learning models~\cite{chen2023towards}.
However, early prior-based methods typically necessitate numerous iterative optimization steps to ascertain the optimal solution, also constraining their practical applicability.
\begin{figure}[!ht]
\centering
%\vspace{-2mm}
\includegraphics[width=1.0\linewidth]{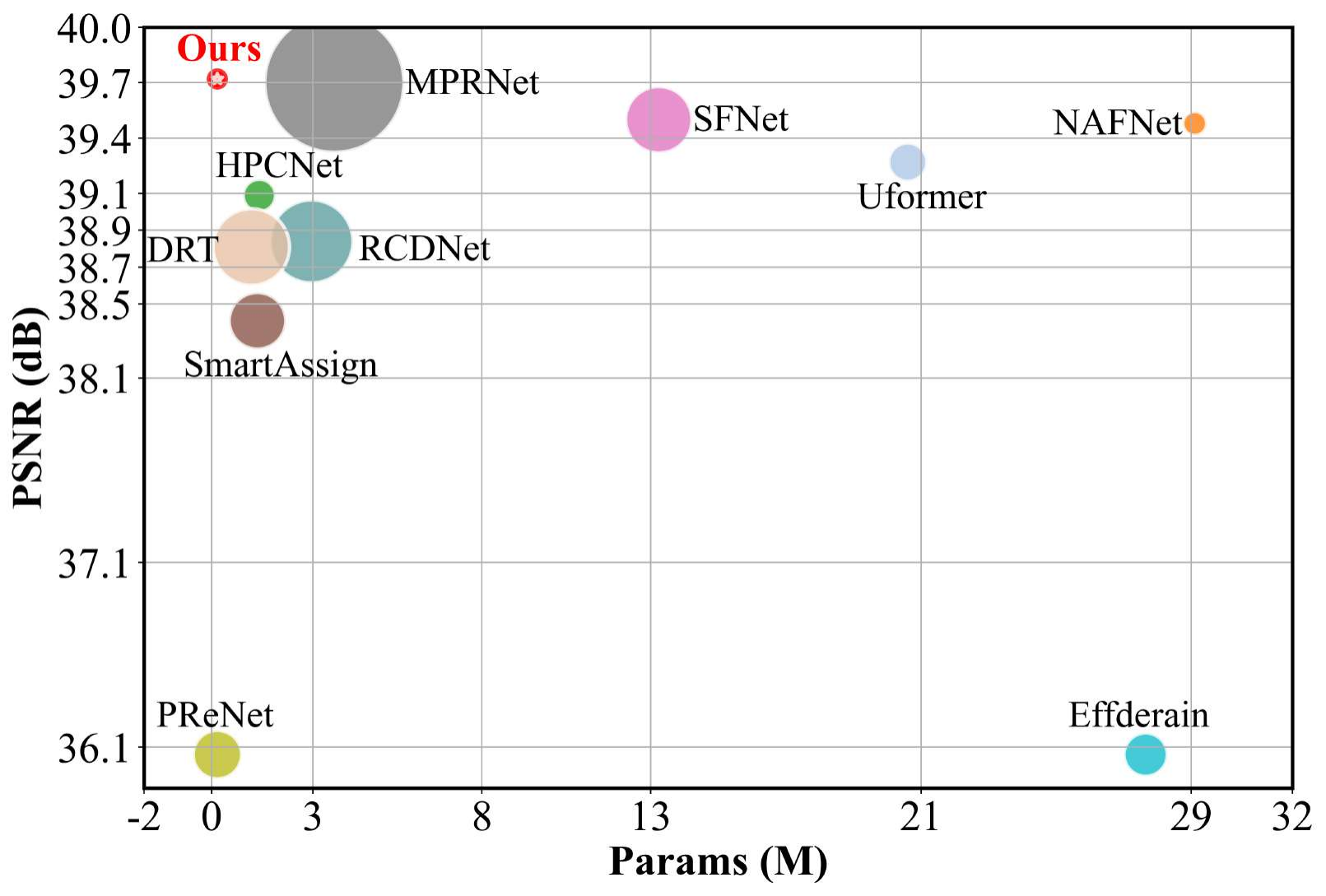}
\vspace{-5mm}
\caption{Model parameters and performance comparison between our proposed ESDNet and other deraining methods on the Rain200L dataset. 
The circle sizes represent the FLOPs of the various methods. 
The results show that the proposed method obtains a superior balance between model complexity and deraining performance.}
\label{fig1}
\vspace{-5mm}
\end{figure}

In the deep learning era, numerous convolutional neural networks (CNNs) based models~\cite{li2018RESCAN,fu2019lpnet,ren2019PReNet,chen2021Unpaired} have been developed as better solutions for image deraining.
The success of these methods greatly stems from their elaborate network architectures, such as multi-scale~\cite{jiang2020mspfn} and  multi-stage~\cite{zamir2021MPRNet}.
However, these networks heavily rely on increasing the depth or complexity of models in order to expand the receptive fields and achieve better performance.

Recently, to facilitate global information aggregation, Transformers~\cite{jie2022IDT,chen2024bidirectional} have been applied to image deraining field and achieved significant advancements as
they can model the non-local information for better image reconstruction.
However, the computation of the scaled dot-product self-attention in Transformers results in quadratic space and time complexity as the number of tokens increases.
To alleviate the computational and memory burden, some efforts have been proposed to reduce model parameters or complexity (FLOPs), such as pruning~\cite{zou2022dreaming}, feature reuse~\cite{fu2019lpnet}, pixel-wise dilation filtering~\cite{guo2021efficientderain}, token sparsification~\cite{chen2023drsformer} and low-rank matrices~\cite{xiong2021nystromformer}. 

\begin{figure}[!t]
\centering
\includegraphics[width=1.0\linewidth]{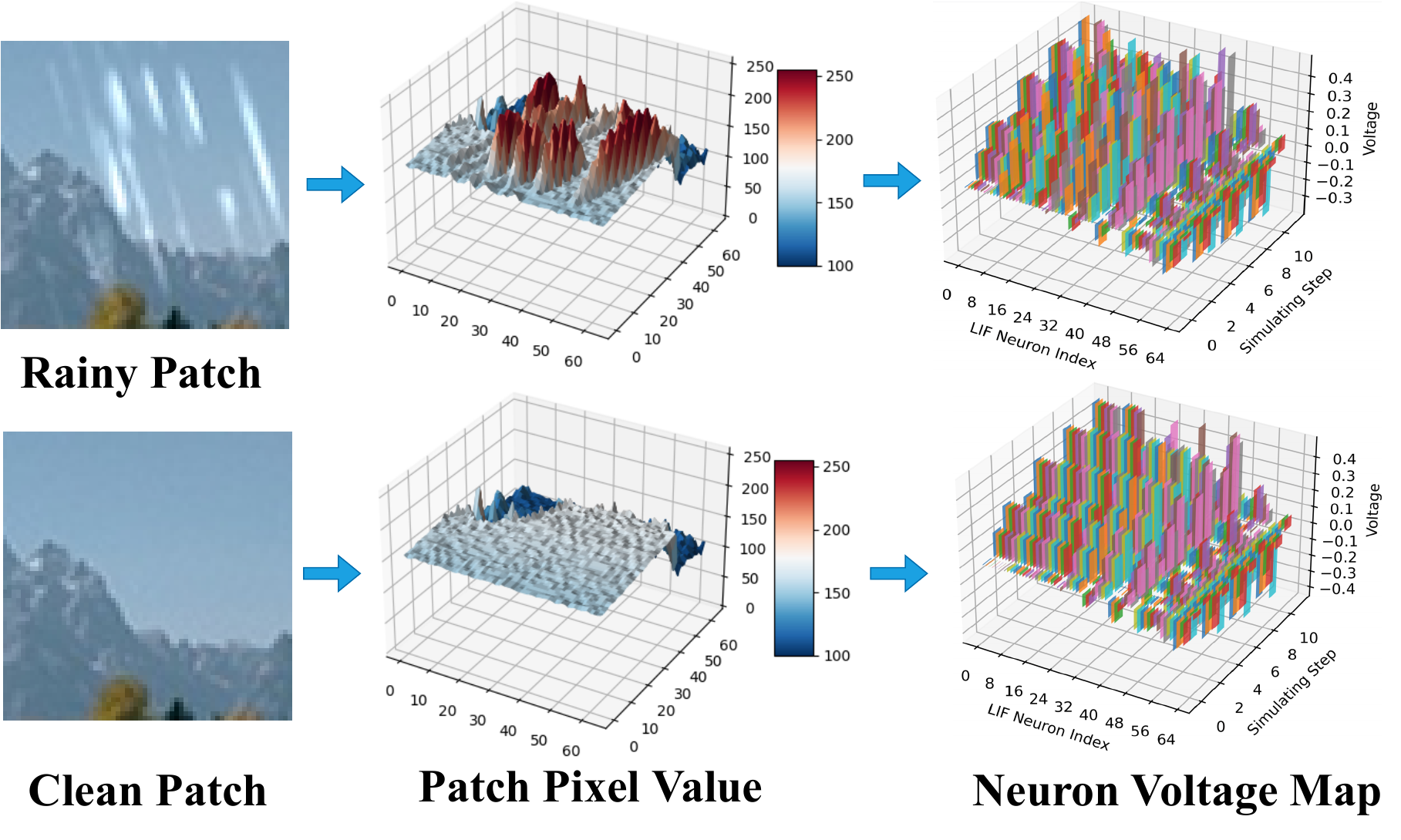}
\vspace{-6mm}
\caption{Visualization of image patch. From the pixel value map, it can be seen that the rain patch is generally higher than the surrounding areas without rain. 
The neuron voltage map represents the voltage heatmap learned by the LIF model. 
For the rain patches, the voltage changes of neurons are more drastic. This indicates that neurons are in an active state with a higher spike rate.}
\vspace{-5mm}
\label{fig2}
\end{figure}

Although these approaches have made initial explorations, achieving a better performance-efficient trade-off remains a major challenge.
Spiking neural networks (SNNs)~\cite{zhou2022spikformer,su2023spike_yolo,qiu2023gatedatten}, as the third generation of neural networks, which aim to simulate the way signals are transmitted in the brain.
Unlike the artificial neural networks (ANNs) that directly learn embedding features from the input image, information in SNNs is represented as binary spike sequences and transmitted by neurons.
Due to strong reflections caused by rain, the values of rain pixels in a rainy image are typically higher than those of their neighboring non-rain pixels.
As shown in Figure~\ref{fig2}, we observe that higher pixel values may result in a more pronounced intensity of spike signals in SNNs because they cause more neurons to fire spikes. 
This naturally arise a question: \textit{whether we can harmonize the characteristics of rain perturbation and SNN for high-quality and low-energy-consuming rain removal?}

In fact, two challenges still need to be addressed when applying the SNN to the image deraining task. 
Firstly, to effectively learn complex and diverse rain streaks, the network needs to possess powerful information representation capabilities. 
Unfortunately, most existing SNN-based methods suffer from information loss caused by discrete binary activation, which may significantly interfere with the subsequent clear image reconstruction.
To enhance the representation capability of SNNs, researchers have explored effective network structures, such as MS-ResNet~\cite{hu2021ms-resnet} and SEW-ResNet~\cite{fang2021sew-resnet}. 
Nevertheless, these networks are not elaborately designed for image deraining task.
Secondly, the non-differentiability of spike signals poses a challenge in using traditional backpropagation training strategies, causing difficulties in training the network.
To address this issue, previous studies have employed the ANN-SNN conversion strategy.
However, this strategy has limitations, as it heavily relies on the performance of the original ANNs and the converted SNN models fail to surpass the performance of the original artificial neural network (ANNs).
A promising approach is to directly train SNNs using surrogate gradients, which can gain better performance with shorter time steps.

To this end, we propose an efficient and effective SNN-based method, called ESDNet, to solve the image deraining problem. 
Specifically, to tackle the weak representation capabilities and scale problems of SNN, we design a spiking residual block (SRB), which consists of a spiking convolution unit (SCU) and a mixed attention unit (MAU). 
SRB first converts the input image into spike signals, then adaptively optimizes the membrane potential by introducing attention weights to adjust spike responses, alleviating information loss caused by discrete binary activation.
Furthermore, to overcome the difficulty of training SNNs, we apply the gradient surrogate strategy to directly train the model rather than the ANN-SNN conversion strategy.
Compared to other ANN-based deraining models, our method exhibits lower energy consumption and superior performance (see Figure~\ref{fig1}).

The main contributions can be summarized as follows:
\begin{itemize}
    \item We propose a deep spike neural network (ESDNet) for single image deraining, and provide a new perspective on developing efficient models for resource-limited devices in real-world applications.
    
    \item We design a spiking residual block, which can effectively alleviate the problem of information loss caused by discrete binary activation and learn multi-scale rain streaks, facilitating high-quality image reconstruction.

    \item Extensive experimental results on single image deraining datasets display that our proposed model can obtain comparable rain removal performance against ANNs-based methods while reducing energy cost by 54\%.
\end{itemize}

\begin{figure*}[!ht]
\centering
\includegraphics[width=1.0\linewidth]{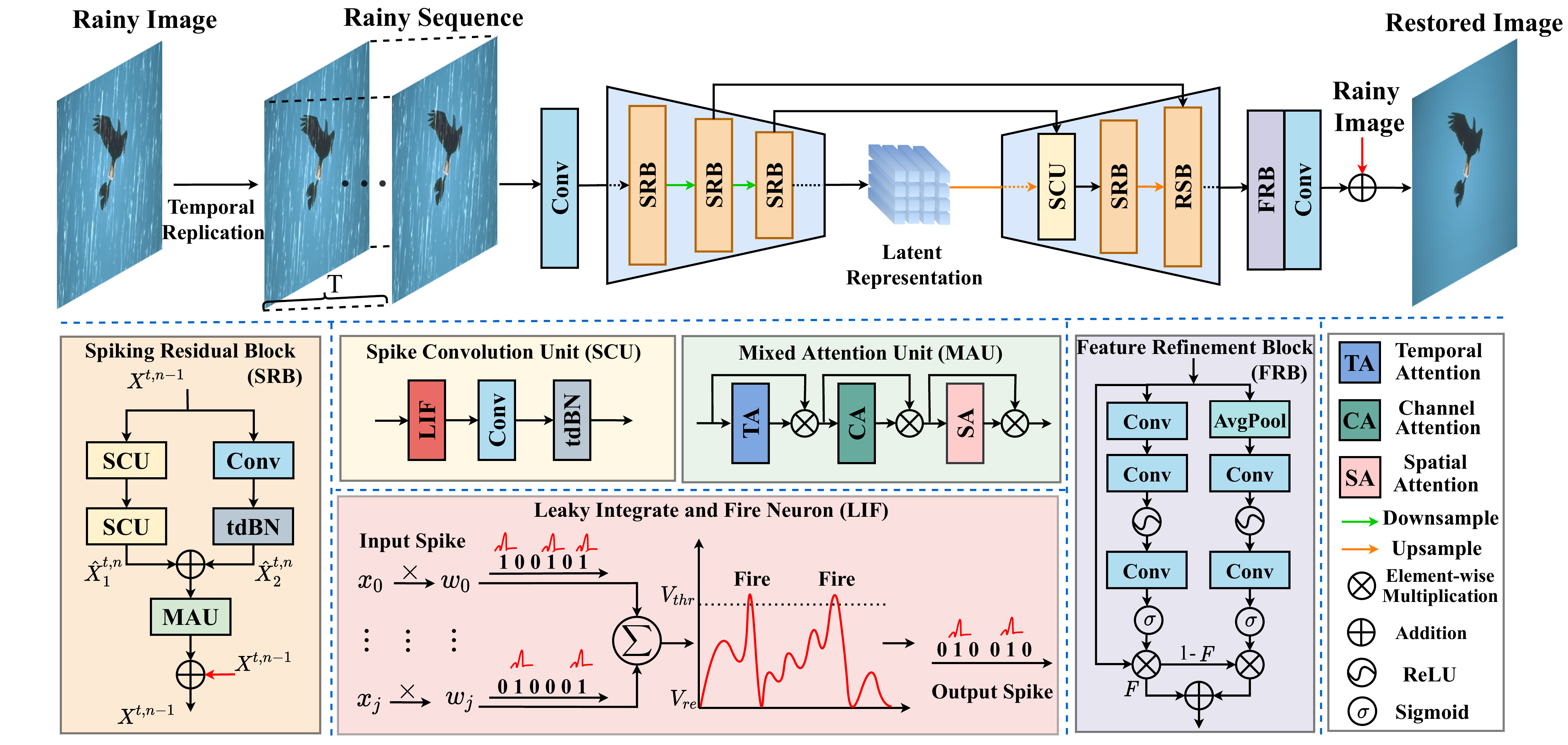}
\vspace{-4mm}
\caption{The architecture of the proposed Efficient Spiking Deraining Network (ESDNet), which takes input rainy images and generates output derained images. 
It mainly contains (1) the Spiking Residual Block (SRB) with Spike Convolution Unit (SCU) and Mixed Attention Unit (MAU), (2) Feature Refinement Block (FRB).
$\operatorname{tdBN}$ refers to threshold-dependent batch normalization.
}
\label{fig3}
\vspace{-4mm}
\end{figure*}
\vspace{-2mm}
\section{Related Work}
\subsection{Single Image Deraining}
Traditional methods for image deraining explored various handcrafted prior knowledge to provide additional constraints for this ill-posed inverse problem.
These methods involved complex optimization problems and only learned the characteristics of rain streaks without considering the underlying clear information~\cite{ding2016ImageDecomposition,luo2015DSCg}. 
Recently, existing most methods used CNN-based frameworks, treating image deraining as a pixel regression task and achieving significant performance gains. 
\cite{jiang2020mspfn},~\cite{zamir2021MPRNet} and~\cite{chen2024rethinking} decomposed the deraining task into multiple subspaces, further acquiring multi-scale information to enhance image reconstruction quality. 

With the successful application of Vision Transformer (ViT) in various advanced visual tasks, the ViT has also been applied to image rain removal, achieving better performance than CNN-based models. 
For example, Restormer~\cite{zamir2022restormer} enhanced representation ability by calculating the similarity of tokens at the channel dimension. 
\cite{jie2022IDT} used a hierarchical architecture of window-based attention to capture global correlations.
Despite these methods achieving impressive reconstruction performance, their heavy computation burden made it difficult to deploy them in practical applications on resource-constrained devices.
 
To improve efficiency, many efforts have been made. 
\cite{fu2019lpnet} constructed a lightweight network based on the Laplace pyramid architecture, which significantly improves the model's efficiency.
\cite{guo2021efficientderain} proposed pixel-wise dilated filtering to predict multi-scale information corresponding to each pixel and developed an efficient rain removal framework.
\cite{ren2019PReNet} designed a deraining network using a recursive computing architecture to accelerate model inference. 
In contrast to CNN-based methods that aimed to improve efficiency through carefully designed architectures, Transformer-based approaches focused more on optimizing attention to enhance computational efficiency. 
Sparse representation~\cite{chen2023drsformer,song2023sfg-idt} and low-rank matrix approximation calculation~\cite{xiong2021nystromformer} were employed to compress and accelerate the inference speed of the model, improving efficiency. 
Although these methods have successfully reduced computational costs, they sacrifice the network's representational ability and cannot effectively explore spatial information, which may affect deraining performances. 
In contrast to these ANN-based rain removal methods, we propose using SNN with lower energy efficiency as a framework to achieve efficient image rain removal.
\vspace{-1mm}
\subsection{Spiking Neural Networks}
Currently, there are two main routes to implementing deep SNNs. 
One approach is the ANN-SNN conversion strategy, which involves converting a well-trained and high-performance ANN into a SNN~\cite{diehl2015fast_ann_relu}. 
In this conversion strategy, the basic idea is to replace the ReLU activation functions in the ANN with spiking neurons to achieve the conversion to an SNN. 
However, this strategy has some inherent limitations. 
Firstly, achieving a conversion close to the performance of the ANN typically requires longer time steps, which leads to higher latency and extra computational consumption~\cite{yao2023attention}. 
Secondly, the performance of the converted SNN is highly dependent on the original ANN and cannot surpass it.
The other promising route is to utilize surrogate gradient functions to unfold the SNN over simulation time steps and directly train the SNN using backpropagation or spike timing dependent plasticity (STDP) \cite{wang2023masked}.
Compared to the ANN-SNN conversion strategy, directly trained SNNs can significantly reduce the simulation time step, making them more attractive in terms of energy efficiency~\cite{kim2020unifying}. 
For example,~\cite{su2023spike_yolo} achieved performance equivalent to an ANN with the same architecture using only 4 time steps. 
\cite{zheng2021tdbn} proposed temporal delay batch normalization (tdBN), which greatly improved the depth of the SNN model. 
\cite{hu2021ms-resnet,yao2023spike_driven_T} applied SNNs to classification tasks and achieved good performance. 
Although some works have applied directly trained deep SNNs to regression tasks such as object detection, there has been little exploration in pixel-level regression tasks such as image deraining.
\vspace{-2mm}
\section{Proposed Method}
In this section, we first offer an overview of the proposed ESDNet.
Then, we introduce the spiking residual block (SRB) and the spike convolution unit (SCU).
Next, we describe the feature refinement block (FRB). 
The training strategy and model optimization are described in the final section.
\vspace{-1mm}
\subsection{Data Preprocessing}
Encoding to generate spike signals globally is a typical approach in SNNs for simulating pixel intensity signals in images.
Considering the spatiotemporal properties of SNNs, we first perform direct encoding on the input rainy image $\mathcal{X}_{t}$ to generate a sequence $\mathbf{X}=\left \{\mathcal{X}_{t}\right \}_{t=1}^T $, \emph{\i.e.,} copying the single degraded image $\mathcal{X}_{t}$ as the input for each time step $t$.
\subsection{Network Architecture}
The proposed network architecture is shown in Figure~\ref{fig3}. 
This network takes a single rainy image as input and adopts an encoder-decoder framework to learn hierarchical representations for capturing information in the network and achieving high-quality deraining output.
First, for the rainy input, the network employs the encoding layer to generate the rainy sequence $\mathbf{X}$ and extract shallow features from the $\mathbf{X}$ by $3\times3$ convolutional layers.
Then, the captured shallow features will be fed into N stacked SRBs to conduct spike transformation, feature extraction, and fusion of rain spike information through SCU and MAU.
For SCU, the Leaky Integrate-and-Fire neurons (LIF) \cite{fangSpikingJelly,deng2022temporal} convert and weight the inputs to generate output spike sequences, which consist of 0 and 1.
The value of the output spike from LIF is 1 when the membrane potential exceeds the threshold, otherwise 0.
Next, MAU can adaptively optimize the membrane potential by applying attention weights, leading to adjusting the spike response in a data-driven manner and learning the distribution of rain streaks.

After the decoder stage, the network adopts a FRB to convert the features extracted from the spike sequence into accurate continuous value representations, while refining the potential representations which are used for restoring clear images \cite{zheng2023ccrdehazing}.
Finally, the network generates high-quality and clear image output from the obtained representations by applying a $3\times3$ convolutional layer.
Furthermore, skip connections are employed to assist model training during the feature extraction and image reconstruction stages.
% \cite{wang2023MSFDNet}
\subsection{Spiking Residual Block}
As previously mentioned, regions degraded with rain perturbation have higher pixel values than those of no rain regions, exhibiting noticeable fluctuations. 
Correspondingly, this may lead to pronounced firing spike rates of neurons in SNNs.
Hence, we develop the spiking residual block (SRB) to learn the spike information from the converted rain streaks, which is displayed in Figure~\ref{fig3}. 
We first adopt SCU to convert the inputs into a discrete spike sequence and mine the rain streak clues included in the spike sequence.
Then, an additional branch comprising solely a convolutional layer and a tdBN layer~\cite{zheng2021tdbn} is constructed to mitigate the significant decline in performance resulting from the inadequate information conveyed by SNNs.
Furthermore, inspired by the previous work~\cite{yao2023attention}, we introduce MAU to augment the spike activity rate of neurons and regulate spike responses, better guiding the process of eliminating rain streaks.
Finally, residual learning is applied to SRB to alleviate the issues of gradient vanishing and explosion.
Assuming that the output of the $n$-th SRB at the $t$-th time step is $\mathbf{X}^{t,n}$, the calculated process of SRB can be depicted as:
\begin{equation}
\begin{split}
\hat{\mathbf{X}}_{1}^{t,n}=&\operatorname{SCU}\left(\operatorname{SCU}\left(\mathbf{X}^{t, n-1}\right)\right),
\\ 
\hat{\mathbf{X}}_{2}^{t,n}=&\operatorname{tdBN}\left(\operatorname{Conv}\left(\mathbf{X}^{t, n-1}\right)\right),
\\
\mathbf{X}^{t,n}=&\operatorname{MAU}\left(\hat{\mathbf{X}}_{1}^{t,n} + \hat{\mathbf{X}}_{2}^{t,n}\right)+\mathbf{X}^{t,n-1},
\end{split}
\end{equation}
$\operatorname{tdBN}$ refers to threshold-dependent batch normalization.
It considers both temporal and spatial dimensions during the normalization process to relieve the gradient explosion.
Profiting from the above designs, SRB can enable adaptively learning the spike fluctuations induced by rain streaks, thereby effectively simulating degradation in diverse regions.
\begin{table*}[!ht]
\resizebox{0.71\width}{!}{
\begin{tabular}{cc|ccccccccccccc}
\hline
\multicolumn{2}{c|}{} & \multicolumn{2}{c}{\textbf{Rain12}}                          & \multicolumn{2}{c}{\textbf{Rain200L}} & \multicolumn{2}{c}{\textbf{Rain200H}}      & \multicolumn{2}{c}{\textbf{Rain1200}} & \multicolumn{2}{c}{\textbf{Average}}       &   &         &                                        \\
\multicolumn{2}{c|}{\multirow{-2}{*}{\textbf{Methods}}}         & \textbf{PSNR$\uparrow$}               & \textbf{SSIM$\uparrow$}                & \textbf{PSNR$\uparrow$}               & \textbf{SSIM$\uparrow$}                & \textbf{PSNR$\uparrow$}               & \textbf{SSIM$\uparrow$}                & \textbf{PSNR$\uparrow$}               & \textbf{SSIM$\uparrow$}                & \textbf{PSNR$\uparrow$}               & \textbf{SSIM$\uparrow$}                & \multirow{-2}{*}{\textbf{Params(M)$\downarrow$}} & \multirow{-2}{*}{\textbf{FLOPs(G)$\downarrow$}} & \multirow{-2}{*}{\textbf{Energy(uJ)$\downarrow$}} \\ \hline
\multicolumn{1}{c|}{}                             & RESCAN      & 35.48                        & 0.9499                        & 33.82                        & 0.9547                        & 26.22                        & 0.8219                        & 32.60                        & 0.9271                        & 32.03                        & 0.9134                        & \textbf{0.149}                                & 32.119                              & 4.014$\times10^5$            \\
\multicolumn{1}{c|}{}                             & PReNet      & 36.32                        & 0.9593                        & 36.06                        & 0.9735                        & 27.18                        & 0.8698                        & 33.39                        & 0.9280                        & 33.24                        & 0.9327                        & 0.168                                & 66.249                              & 8.281$\times10^5$            \\
\multicolumn{1}{c|}{}                             & RCDNet      & 37.67                        & 0.9612                        & 38.84                        & 0.9854                        & 28.98                        & 0.8889                        & 32.68                        & 0.9189                        & 34.54                        & 0.9386                        & 2.958                                & 194.501                             & 2.431$\times10^6$            \\
\multicolumn{1}{c|}{} & MPRNet  & 37.55 & 0.9665 
&\underline{39.82} &\underline{0.9863}
&\underline{29.94}  & 0.8999  & 34.50    & 0.9369                 &\underline{35.46} &\underline{0.9474} 
& 3.637  & 548.651   & 6.858$\times10^6$   \\
\multicolumn{1}{c|}{}   & Effderain    & 36.12   & 0.9588   & 36.06   & 0.9731 
& 26.11   & 0.8341      & 32.85        & 0.9147  & 32.78    & 0.9202   & 27.654        & 52.915 & 6.614$\times10^5$            \\
\multicolumn{1}{c|}{}  & NAFNet  & 37.53    & 0.9639   & 39.48   & 0.9847        &29.19    & 0.8880     & 34.69   & 0.9367   & 35.22    & 0.9433  & 29.102             &\underline{16.064}    &\underline{2.008$\times10^5$}            \\
\multicolumn{1}{c|}{\multirow{-8}{*}{\textbf{C}}} & SFNet   & 36.11   & 0.9503        & 39.50   & 0.9850    & 29.75    &\underline{0.9008}             &\underline{34.51} &\underline{0.9383} 
& 34.97  & 0.9437     & 13.234    & 124.439     & 1.555$\times10^6$     
                                        \\ \hline
\multicolumn{1}{c|}{} & DRT   &\underline{37.74} 
&\underline{0.9674} & 38.81  & 0.9830  & 28.67                   & 0.8796   & 33.88   & 0.9283   & 34.78  & 0.9395 & 1.176  & 166.045                  & 2.075$\times10^6$            \\
\multicolumn{1}{c|}{} & ELFformer   & 35.06  & 0.9420  & 38.85  & 0.9800              & 28.93   & 0.8852  & 33.54   & 0.9360   & 34.09  
& 0.9358  & 1.221   & 23.667  & 2.958$\times10^5$            
                            \\
\multicolumn{1}{c|}{}  & HPCNet      & 36.84       & 0.9639     & 39.14               & 0.9847    & 29.17   &0.8962        & 34.46       & 0.9378     & 34.90               & 0.9378    & 1.411   & 29.540       & 3.692$\times10^5$            
\\
\multicolumn{1}{c|}{\multirow{-4}{*}{\textbf{T}}} & SmartAssign & 36.87               & 0.9618     & 38.41    & 0.9814   & 27.71   & 0.8536    & 33.11                    
& 0.9154     & 34.03    & 0.9281   & 1.359   & 90.386    & 1.129$\times10^6$            \\ \hline
\multicolumn{1}{c|}{\textbf{S}}               & Ours     &\textbf{37.82} & \textbf{0.9681} & \textbf{39.85} &\textbf{0.9869} & \textbf{30.01} & \textbf{0.9132} & \textbf{34.52} &\textbf{0.9388} & \textbf{35.55} &\textbf{0.9518} &\underline{0.165} &\textbf{7.320}  & \textbf{9.150$\times 10^4$} 
\\ \hline           
\end{tabular}
}
\caption{Comparison of quantitative results on four synthetic benchmarks. 
\textbf{C}, \textbf{T} and \textbf{S} refer to the CNN-based, Transformer-based and SNN-based baselines, respectively. 
\textbf{Blod} and \underline{underline} indicate the best and second-best results, respectively.}
\label{tab1}
\vspace{-2mm}
\end{table*}
\subsection{Spike Convolution Unit}
Compared to ANNs, which employ continuous decimal values for information propagation and calculation, SNNs exploit discrete binary spike values for communication.
Spiking neurons can convert received continuous values into discrete spike sequences, encoding information with temporal precision.
In this paper, we select LIF neurons to convert input signals into spike sequences, as they better balance biological characteristics and computational complexity~\cite{guo2023rmp}.
The explicit dynamic equation for LIF neurons can be expressed mathematically as:
\begin{equation}
\left\{\begin{array}{l}
\boldsymbol{H}^{t, n}=\boldsymbol{U}^{t-1, n}+\frac{1}{\tau}\left ( \boldsymbol{X}^{t, n}-(\boldsymbol{U}^{t-1, n}-V_{\text {reset}})\right ) \\
\boldsymbol{S}^{t, n}=\Theta \left(\boldsymbol{H}^{t, n}-V_{thr}\right) \\
\boldsymbol{U}^{t, n}=\left(\beta \boldsymbol{H}^{t, n}\right) \odot\left(\mathbf{1}-\boldsymbol{S}^{t, n}\right)+V_{\text {reset}} \boldsymbol{S}^{t, n},
\end{array}\right.
\label{eq2}
\end{equation}
where $t$ and $n$ represent the $t$-th time step and the $n$-th layer.
$\boldsymbol{H}^{t, n}$ represents membrane potential, which is generated by a combination of time input $\boldsymbol{U}^{t-1, n}$ and spatial input $\boldsymbol{X}^{t, n}$.
$\tau$ is the membrane time constant.
Among them, $V_{thr}$ represents the activation threshold, used to determine whether to output $\boldsymbol{S}$ or maintain it at 0.
$\Theta(\cdot)$ denotes the step function.
When $x\ge 0$, $\Theta(x)=1$, otherwise $\Theta(x)=0$.
$V_{\text{reset}}$ denotes the potential reset after $\boldsymbol{S}$ activates the output.
% 
% $\beta=e^{-\frac{d t}{\tau}}<1$ 
$\beta$ represents the decay factor and $\odot$ indicates the element-wise multiplication.

In Eq.~\ref{eq2}, the spatial representation $\boldsymbol{X}^{t, n+1}$ of rain streaks is captured from the output spike sequence $\boldsymbol{U}^{t, n}$ by convolution operators.
This calculation process can be defined as:
\begin{equation}
\boldsymbol{X}^{t, n+1}=\operatorname{tdBN}\left(\operatorname{Conv}\left( \boldsymbol{S}^{t, n}\right)\right),
\end{equation}
where $\boldsymbol{S}^{t, n}$ denotes a spike tensor, which only included 0 and 1.
$\operatorname{tdBN}$ refers to threshold-dependent batch normalization.
Due to space limitations, specific descriptions of MAU are included in the supplementary materials.

\subsection{Feature Refinement Block}
To transform discrete pulse sequences into continuous pixel values, a typical approach is to apply mean sampling in the time dimension and then scale it to the input size. 
Nevertheless, this sampling process may result in the loss of some crucial structural elements, which could affect the final image restoration quality.
To address this issue, we introduce a FRB that can adaptively aggregate sampled features and mitigate information loss. 
Given the input spike sequence $\boldsymbol{Y}$, the calculation process of FRB can be presented as:
\begin{equation}
\begin{split}
\hat{\boldsymbol{F}}&=\sigma\left(\operatorname{Conv}\left(\operatorname{\phi }\left(\operatorname{Conv}(\operatorname{GAP}(\boldsymbol{Y}))\right)\right)\right),
\\
\boldsymbol{F}&=\sigma\left(\operatorname{Conv}\left(\operatorname{\phi }\left(\operatorname{Conv}(\operatorname{Conv}(\boldsymbol{Y}))\right)\right)\right),
\\
\Tilde{\boldsymbol{F}}&=\boldsymbol{Y}\odot\boldsymbol{F}+(1-\boldsymbol{F})\odot\hat{\boldsymbol{F}},
\end{split}
\end{equation}
where $\Tilde{\boldsymbol{F}}$ refers to the final output feature and $\operatorname{GAP}(\cdot)$ denotes the global average pooling operation.
$\sigma(\cdot)$ and $\phi(\cdot)$ indicate the Sigmoid and ReLU functions.
Subsequently, the output feature of the FRB will be fed into a $3 \times3$ convolutional layer to map it back to the original input resolution.
By incorporating FRB into our proposed model, the network can better learn and refine the latent representation of clear images for restoring fine structural and textural details in images.
\subsection{Training Strategies}
In recent years, backpropagation-based training schemes have achieved significant progress in facilitating the training of SNNs, showing considerable superiority over transformation-based strategies.
Since the non-differentiable nature of binary activations in SNN neurons, most existing methods use gradient-proxy functions to implement backpropagation training. 
Inspired by the previous wisdom, the Sigmoid function serves as the gradient-surrogate function learning to train SNNs. 
It can effectively handle binary neuron outputs and enable gradient propagation during backpropagation.
The gradient surrogate function can be defined as:
\begin{equation}
\begin{split}
\sigma(x)=&\frac{1}{1+e^{-\alpha x}}, \\
\sigma^{\prime}(x)=\alpha \cdot &\sigma(x) \cdot(1-\sigma(x)),
\end{split}
\end{equation}
where $\alpha$ refers to a hyper-parameter, adopted to adjust the gradient of the surrogate function. 
The larger the value of $\alpha$, the greater the gradient of the function.

\subsection{Model Optimization}
To better recover more similar structure details, we apply SSIM loss as the reconstruction loss and train the model by minimizing it.
The loss function can be depicted as:
\begin{equation}
\mathcal{L}_{ssim}=1-\operatorname{SSIM}(\mathcal{S}(\mathcal{X}), \mathcal{Y}),
\end{equation}
where $\mathcal{S}(\cdot)$ is the proposed ESDNet, $\mathcal{X}$ and $\mathcal{Y}$ are the input rainy image and its corresponding ground-truth, respectively.

\begin{table*}[!t]
\resizebox{0.81\width}{!}{
\begin{tabular}{ccccccccccccc}
\hline
\multicolumn{1}{c}{Methods} & RESCAN  & PReNet & RCDNet & MPRNet & Effderain & NAFNet & SFNet  & DRT                          & ELFformer                     & HPCNet & SmartAssign & Ours                          \\ \hline
% Datasets                     & \multicolumn{13}{c}{RW-Data}                                                                       \\ 
PIQE$\downarrow$                        & 18.364  & 18.309 & 19.692 & 18.061 & 18.400   & 18.106 & 18.124 & 18.392                       &\underline{18.037} & 18.287 & 18.316      & \textbf{17.959} \\ \hline
MetaIQA$\uparrow$    & 0.419    & 0.420  & 0.421  &\underline{0.426}  & 0.421    & 0.420  & 0.423  & 0.419 & 0.422                         & \textbf{0.426}  & 0.417       &0.423  \\ \hline
\end{tabular}
}
\vspace{-1mm}
\caption{Comparison of PIQE/MetaIQA scores on the RW-Data.
\textbf{Blod} and \underline{underline} indicate the best and second-best results, respectively.
}
\vspace{-1mm}
\label{tab2}
\end{table*}

\begin{figure*}[!t]
\centering
\begin{minipage}{0.22\textwidth}
    \centering
    \includegraphics[width=\textwidth]{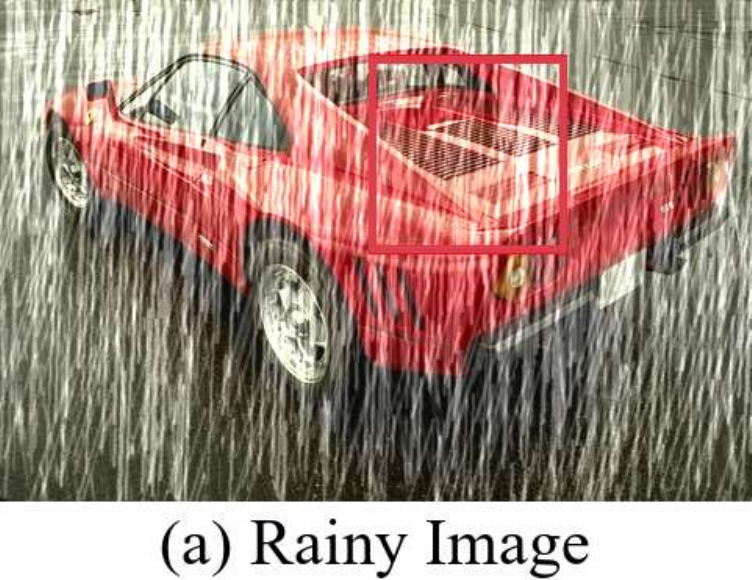}
\end{minipage}%
\hfill 
\begin{minipage}{0.77\textwidth}
\centering
    \includegraphics[width=\textwidth]{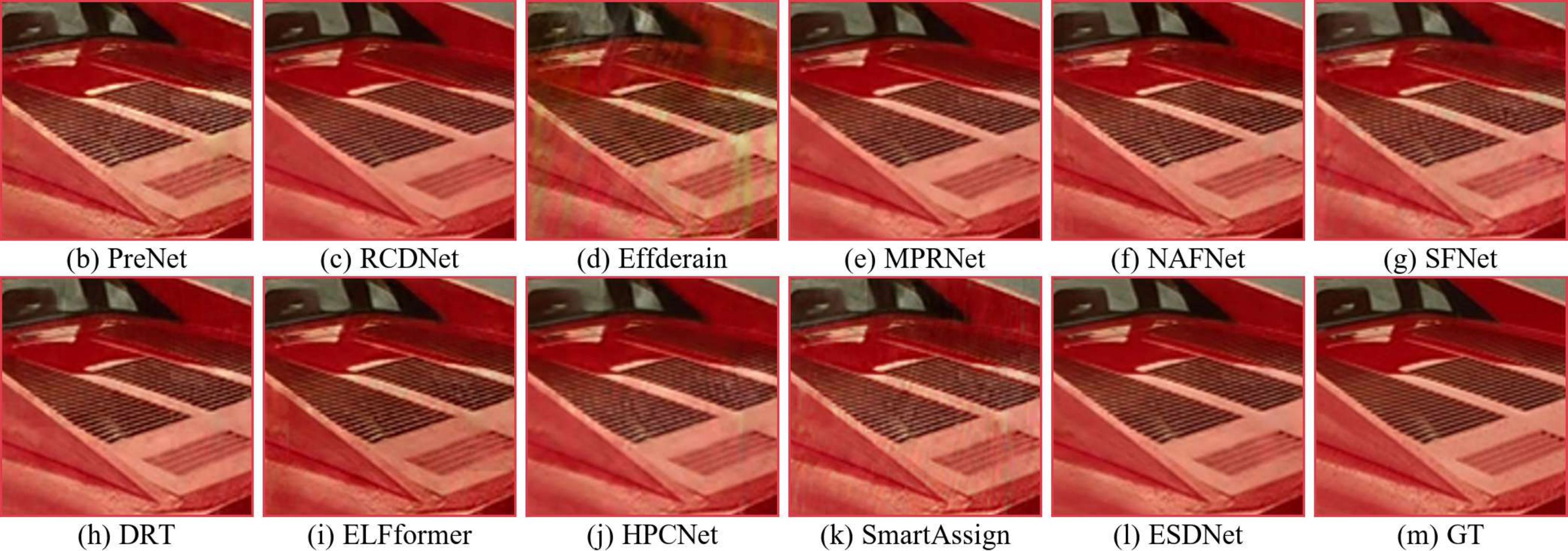}
\end{minipage}
\vspace{-1mm}
\caption{Visual comparison of the Rain200H dataset. Best viewed by zooming in on the figures on high-resolution displays.
}
\vspace{-4mm}
\label{fig4}
\end{figure*}

\section{Experiments}
In this section, we conduct comprehensive experiments on commonly used benchmark datasets to evaluate the effectiveness of the proposed method. 
The experiments involve rigorous testing and analysis, aiming to provide a thorough assessment of the performance. 
More results are included in the supplemental material. The training code and test models will be available to the public.

\subsection{Experimental Setup}\label{sec4.1}
{\flushleft\textbf{Datasets}.} Since the original comparison methods are trained on different datasets, we retrained all models on the four publicly available datasets (Rain12~\cite{li2016GMM}, Rain200L~\cite{yang2017JORDER}, Rain200H~\cite{yang2017JORDER}, Rain1200~\cite{zhang2018diddata}) to ensure a fair comparison of all methods. 
The Rain12 dataset consists of 12 synthetic rain images that were used for testing. 
The Rain200L and Rain200H datasets contain 1800 synthetic rain images for training, along with 200 images designated for testing. 
The Rain1200 dataset includes 12000 training images and 1200 test images, which have varying rain directions and density levels. 
Furthermore, we exploit the RW-Data~\cite{zhang2019ID-GAN}, a real-world rain benchmark, consisting of 185 real-world rainy images.
{\flushleft\textbf{Evaluation Metrics}.} In light of practical considerations, we select two common evaluation metrics, \emph{e.g.,} Peak Signal to Noise Ratio (PSNR)~\cite{hore2010PSNR} and Structural Similarity (SSIM)~\cite{wang2004SSIM}, to perform a quantitative comparison of the synthesized rainy dataset. 
Following the previous approaches~\cite{chen2023drsformer}, we compute these metrics in the Y channel of the YCbCr space. 
For the real-world rain dataset, we adopt non-reference image quality evaluation metrics such as Perception-based Image Quality Evaluator (PIQE)~\cite{venkatanath2015PIQE} and Meta-learning-based Image Quality Assessment (MetaIQA)~\cite{zhu2020metaiqa} for assessing performance.
{\flushleft\textbf{Comparison Methods}.} 
We compare our proposed method with various image rain removal baselines, including seven CNN-based methods (\emph{i.e.}, RESCAN~\cite{li2018RESCAN}, PreNet~\cite{ren2019PReNet}, RCDNet~\cite{wang2020RCDNet}, MPRNet~\cite{zamir2021MPRNet}, Effderain~\cite{guo2021efficientderain}, NAFNet \cite{chen2022nafnet}, and SFNet~\cite{cui2022SFNet}) and four transformer-based networks (\emph{i.e.}, DRT~\cite{liang2022drt}, ELFormer ~\cite{jiang2022elformer}, HPCNet~\cite{wang2023hpcnet}, and SmartAssign ~\cite{wang2023smartassign}).
For the recent representative method SmartAssign, we encounter a lack of available code. 
Hence, we recreate the corresponding network according to the provided information in their paper and retrain it.
For other competitive methods, if the authors provide pre-trained models available, we utilize their online code to evaluate performance, otherwise, we will retrain them using the same settings as ESDNet to ensure a fair comparison.
{\flushleft\textbf{Implementation Details.}} 
During the training process, we conduct the proposed network in the PyTorch framework with an Adam optimizer and a batch size of 12. 
We set the learning rate to $1\times 10^{-3}$ and apply the cosine annealing strategy \cite{song2023learning} to steadily decrease the final learning rate to $1 \times10^{-7}$. 
For Rain200L, Rain200H, and Rain1200 datasets, we train the model by 1000 epochs. 
We set the stacking numbers of SRB to [4,4,8] in the encoder stage and [2,2] in the decoder stage.
For the $\alpha$ of the gradient proxy function, it is set to 4 according to~\cite{su2023spike_yolo}.
All experiments are executed on an NVIDIA GeForce RTX 3080Ti GPU (12G). 
To ensure a fair comparison, for all retrained methods, we uniformly cut the data into 64 patch sizes for training, and apply the sliding window slicing strategy for testing.

\subsection{Experimental Results}
{\flushleft\textbf{Synthetic Datasets.}} Table~\ref{tab1} provides a comprehensive comparison between our proposed method and 12 representative and competitive training methods. 
It is evident that the incorporation of spike convolution units in our ESDNet significantly improves the performance in terms of PSNR and SSIM values compared to all other baselines. 
Notably, our approach achieves more appealing results on the Rain200H benchmark, surpassing the recent CNN-based method SFNet by 0.26 dB in PSNR.
The performance improvement, when compared to existing ANNs-based deraining methods, shows that our framework offers a fresh perspective for deep image training architectures.
Furthermore, Figure~\ref{fig3} presents the qualitative evaluation results on the Rain200H benchmark. 
The visual comparison indicates that our method exhibits enhanced contrast and reduced color distortion in comparison to other approaches, corroborating the quantitative findings.
Specifically, Effderain and DRT exhibit noticeable rain artifacts in Figure~\ref{fig3}, while also tending to produce overly smooth results. 
In contrast, our model preserves more details and achieves superior perceptual quality.
\begin{figure*}[!t]
\centering
\includegraphics[width=1.0\linewidth]{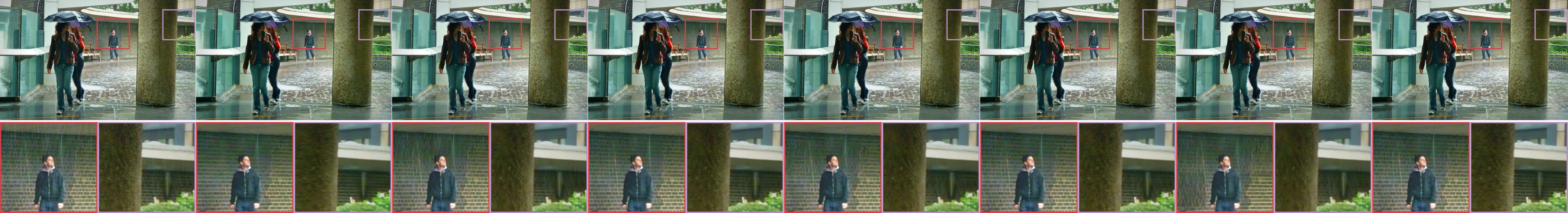}
\includegraphics[width=1.0\linewidth]{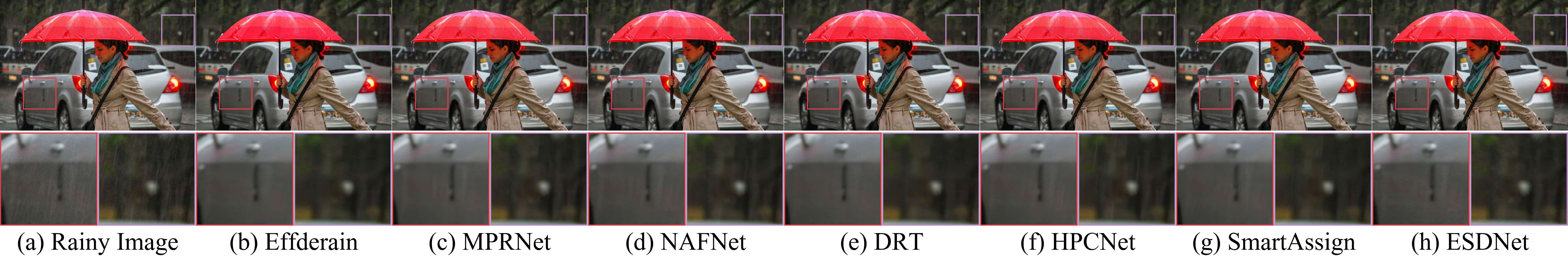}
\vspace{-3mm}
\caption{Visual comparison on the RW-Data. Best viewed by zooming in the figures on high-resolution displays.}
\label{fig5}
\vspace{-3mm}
\end{figure*}
{\flushleft\textbf{Real-world Datasets.}} We conduct additional experiments using the RW-Data benchmark dataset. The obtained quantitative results are presented in Table \ref{tab2}. 
Analysis of these results reveals that our network achieves lower PIQE values compared to other comparative models. 
This indicates that in real rain scenarios, our network exhibits superior output quality with clearer content and enhanced perceptual quality.
To further evaluate the qualitative performance, we implement a quality comparison analysis as shown in Figure \ref{fig4}. 
The results demonstrate that most models exhibit sensitivity to spatial long rain streaks, which often leads to noticeable rain effects in their outputs. 
In contrast, our network effectively eliminates a significant portion of rain disturbances, resulting in visually pleasing restoration effects. 
Overall, the results suggest that our network outperforms existing models in terms of output quality, perceptual clarity, and the ability to handle rain disturbances in real scenarios.

{\flushleft\textbf{Model Effciency}.} In ANNs, each operation involves the multiply-accumulate operations (MAC) and floating-point operations (FLOPs) are used to estimate computational burden.
SNNs exhibit energy-saving properties in neuromorphic hardware, as neurons only participate in accumulated operations (AC) during spikes and can complete AC through approximately the same number of synaptic operations (SOPs).
In this work, we exploit 4 time steps to estimate energy costs. 
It is observed that the first convolutional encoding layer, RB block, and last convolutional encoding layer require floating-point operations due to they enjoy the matrix multiplication. 
The other layers are calculated through SOPs (synaptic operations).
Following the previous works \cite{guo2023ternary}, SOPs can be computed through $s\times T\times A$, where s represents the average sparsity, $T$ is the time step, and $A$ is the additive number in the artificial neural network. 
For the binary SNNs, the sparsity is 16.42\%.
It is noted that SOPs can be triggered once when the sign function output is 1 due to the membrane potential exceeding the threshold. One sign (using energy per spike to calculate) needs 3.7pJ.
Furthermore, the sign function only exists in LIF, and the number of LIF is limited compared to convolution operations.
According to \cite{hu2021spikingresnet}, we compute the energy as follows: a FLOP requires 12.5pJ, an SOP requires 77fJ, and a Sign (calculated using energy per spike) requires 3.7pJ.
\begin{table}[!ht]
\resizebox{0.90\width}{!}{
\begin{tabular}{ccccc}
\hline
\multirow{2}{*}{Model} & \multirow{2}{*}{FLOPs (G)} & \multirow{2}{*}{Energy (uJ)} & \multirow{2}{*}{PSNR} & \multirow{2}{*}{SSIM} \\
                                &                                    &                                      &                                &                                \\ \hline
ANN-ResNet                      & 83.453                            &1.043$\times 10^6$                                      & 28.79                         & 0.8942                          \\
MS-ResNet                       & 6.970                             &8.730$\times 10^4$                                      & 25.20                         & 0.8230                         \\
SEW-ResNet                      & 6.998                             &8.765$\times 10^4$                                      & 26.61                         & 0.8613                         \\ \hline
SRB                            & 7.320                             &9.150$\times 10^4$                                      & \multicolumn{1}{c}{29.31}    & 0.9033                         \\ \hline
\end{tabular}
}
\caption{Ablation studies for different residual blocks in the model.
}
\label{tab3}
\end{table}

\begin{table}[!ht]
\resizebox{0.95\width}{!}{
\begin{tabular}{ccccc}
\hline
\multirow{2}{*}{Time steps} & \multirow{2}{*}{FLOPs (G)} & \multirow{2}{*}{Energy (uJ)} & \multirow{2}{*}{PSNR} & \multirow{2}{*}{SSIM} \\
                       &                           &                             &                       &                       \\ \hline
1                      & 7.023                    &8.779$\times 10^4$                             & 28.69                & 0.8645                \\
2                      & 7.150                    & 8.938$\times 10^4$                             & 28.81                & 0.8963                \\
4                      & 7.320                    &9.150$\times 10^4$                              & 29.31                & 0.9033                \\ 
6                      & 7.490                    &9.366$\times 10^4$                              & 29.40                & 0.9040                \\ \hline
\end{tabular}
}
\vspace{-2mm}
\caption{Impact of different time steps on Rain200H dataset.
}
\label{tab4}
\vspace{-2mm}
\end{table}

The results are displayed in Table \ref{tab1}. 
It can be observed that our network has achieved performance comparable to the recent ANNs-based deraining approaches, and has lower energy costs.
In addition, we compared the computational complexity of various deep derivation transformer-based methods, including the number of trainable parameters and FLOPs on $256\times 256$ images, as reported in Figure~\ref{fig1}.

\subsection{Ablation Studies}
We conduct the ablation experiments to better understand the impact of different residual blocks, time steps, and model components on performance.
Unless otherwise specified, all ablation experiments are conducted on the Rain200H dataset using the experimental setup described in Section~\ref{sec4.1}.

{\flushleft{\textbf{Effectiveness of Different Residual Blocks}.}} To investigate the impact of different residual modules on model performance, Table~\ref{tab3} presents the PSNR/SSIM values of the corresponding models.
From the experimental results, we note that our SRB obtains a more attractive performance compared to the other two spiking residual blocks, thanks to the additional branches in SRB and the information compensation brought by MAU.
\begin{table}[!ht]
\resizebox{0.97\width}{!}{
\begin{tabular}{ccccccc}
\hline
Model & BN & tdBN                 & MAU & FRB                  & PSNR   & SSIM   \\ \hline
(a)   & $ \checkmark$  &                      &     &                      & 28.26 & 0.8779 \\
(b)   &    & $ \checkmark$                    &     &                      & 28.49 & 0.8724 \\
(c)   & $\checkmark$  & \multicolumn{1}{l}{} & $ \checkmark$   & \multicolumn{1}{l}{} & 29.07 & 0.8984 \\
(d)   &    & $\checkmark$                    &     & $\checkmark$                   & 28.59 & 0.8931 \\
(e)   &    & $ \checkmark$                    & $ \checkmark$   &                      & 28.78 & 0.8968 \\
(f)   &    & $ \checkmark$                    & $ \checkmark$   & $ \checkmark$                    & 29.31 & 0.9033 \\ \hline
\end{tabular}
}
\vspace{-2mm}
\caption{Ablation studies for different designs in the model.}
\label{tab5}
\vspace{-2mm}
\end{table}
By introducing a lower computational burden alone, the model performance is significantly improved.

{\flushleft{\textbf{Effectiveness of Different Time Steps}}.} The performance results of the model with different time steps are reported in Table~\ref{tab4}.
According to the experimental results, it can be found that the longer the time step, the higher the quality of image restoration.
This indicates that higher time steps can enhance the model to better obtain feature representations.
Similarly, this also brings about a significant increase in energy costs.
To better balance performance and efficiency, we chose a time step of 4 as the default time step in this work.

{\flushleft{\textbf{Effectiveness of Different Designs}}.} 
To evaluate the effectiveness of the model designs, we conduct experiments based on different model variables in Table~\ref{tab5}.
Compared to the baseline model (a), model (b) provides additional performance advantages due to the consideration of temporal correlation.
In addition, we note a significant decrease in performance when MAU is not applied, indicating that MAU can effectively enhance the spike activity rate of neurons and regulate spike responses.

\section{Concluding Remarks}
This paper presents an image deraining approach using spiking neural networks. 
The proposed network learns rain streak information by leveraging the spike convolution unit.
Initially, continuous pixel values are transformed into spike sequences by LIF neurons.
Convolution operations are then applied to learn the spike information associated with rain streaks. 
To enhance the accuracy of rain streak removal, a mixed attention unit is introduced, which adjusts spike responses based on the input data, thereby facilitating the extraction of rain streak clues. 
In addition, feature refinement blocks are incorporated to further improve the model representation for gaining high-quality images. 
The efficacy of the proposed method is demonstrated through extensive experiments conducted on synthetic and real rain datasets.

\bibliographystyle{named}
\bibliography{ijcai24}
\end{document}